\let\accentvec\vec
\documentclass[runningheads,a4paper]{llncs}

\let\vec\accentvec
\usepackage{amssymb}
\usepackage{amsmath}
\usepackage{bbm}
\usepackage{xfrac}
\usepackage{graphicx}
\usepackage{cite}
\usepackage{multirow}
\usepackage{booktabs}
\usepackage{tabularx}
\usepackage{array}
\usepackage{epstopdf}
\newcolumntype{+}{>{\global \let \currentrowstyle \relax }}
\newcolumntype{^}{>{\currentrowstyle }}

\usepackage{url}
\urldef{\mailsa}\path| *@*|

\newcolumntype{Y}{>{\centering\arraybackslash}X}
\begin{document}
\mainmatter 
\title{Real-time 3D Tracking of Articulated Tools for Robotic Surgery}
\titlerunning{Real-time 3D Tracking of Articulated Tools for Robotic Surgery}
%
%
\author{Menglong Ye, Lin Zhang, Stamatia Giannarou and Guang-Zhong Yang
}
\authorrunning{M. Ye et al.}
\institute{The Hamlyn Centre for Robotic Surgery, Imperial College London, UK\\
\url{menglong.ye11@imperial.ac.uk}
}
%
%
\maketitle

\begin{abstract}

In robotic surgery, tool tracking is important for providing safe tool-tissue interaction and facilitating surgical skills assessment. Despite recent advances in tool tracking, existing approaches are faced with major difficulties in real-time tracking of articulated tools. Most algorithms are tailored for offline processing with pre-recorded videos. In this paper, we propose a real-time 3D tracking method for articulated tools in robotic surgery. The proposed method is based on the CAD model of the tools as well as robot kinematics to generate online part-based templates for efficient 2D matching and 3D pose estimation. A robust verification approach is incorporated to reject outliers in 2D detections, which is then followed by fusing inliers with robot kinematic readings for 3D pose estimation of the tool. The proposed method has been validated with phantom data, as well as \emph{ex vivo} and \emph{in vivo} experiments. The results derived clearly demonstrate the performance advantage of the proposed method when compared to the state-of-the-art.

\end{abstract}

\section{Introduction}
Recent advances in surgical robots have significantly improved the dexterity of the surgeons, along with enhanced 3D vision and motion scaling. Surgical robots such as the da Vinci\textsuperscript{\textregistered} (Intuitive Surgical, Inc. CA) platform, can allow the augmentation of preoperative data to enhance the intraoperative surgical guidance. In robotic surgery, tracking of surgical tools is an important task for applications such as safe tool-tissue interaction and surgical skills assessment.

In the last decade, many approaches for surgical tool tracking have been proposed. The majority of these methods have focused on the tracking of laparoscopic rigid tools, including using template matching \cite{Sznitman2012} and combining colour-segmentation with prior geometrical tool models \cite{Wolf2011}. In \cite{Allan2015}, the 3D poses of rigid robotic tools were estimated by combining random forests with level-sets segmentation. More recently, tracking of articulated tools has also attracted a lot of interest. For example, Pezzementi et al. \cite{Pezzementi2009} tracked articulated tools based on an offline synthetic model using colour and texture features. The CAD model of a robotic tool was used by Reiter et al. \cite{Reiter2012} to generate virtual templates using the robot kinematics. However, thousands of templates were created by configuring the original tool kinematics, leading to time-demanding rendering and template matching. In \cite{Sznitman2014}, boosted trees were used to learn predefined parts of surgical tools. Similarly, regression forests have been employed in \cite{Rieke2015} to estimate the 2D pose of articulated tools. In \cite{Reiter2014}, the 3D locations of robotic tools estimated with offline trained random forests, were fused with robot kinematics to recover the 3D poses of the tools. Whilst there has been significant progress on surgical tool detection and tracking, none of the existing approaches have thus far achieved real-time 3D tracking of articulated robotic tools. 

In this paper, we propose a framework for real-time 3D tracking of articulated tools in robotic surgery. Similar to \cite{Reiter2012}, CAD models have been used to generate virtual tools and their contour templates are extracted online, based on the kinematic readings of the robot. In our work, the tool detection on the real camera image is performed via matching the individual parts of the tools rather than the whole instrument. This enables our method to deal with the changing pose of the tools due to articulated motion. Another novel aspect of the proposed framework is the robust verification approach based on 2D geometrical context, which is used to reject outlier template matches of the tool parts. The inlier 2D detections are then used for 3D pose estimation via the Extended Kalman Filter (EKF). Experiments have been conducted on phantom, \emph{ex vivo} and \emph{in vivo} video data, and the results verify that our approach outperforms the state-of-the-art. 
\begin{figure}[t]
\centering
\includegraphics[width=\linewidth]{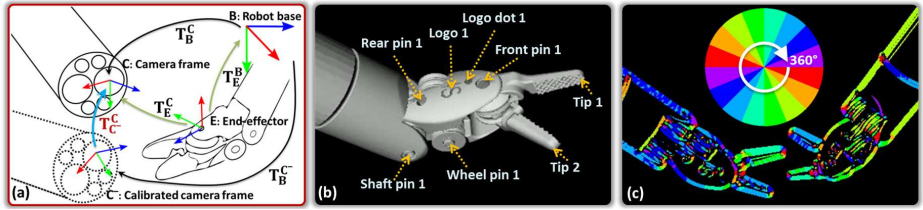}
\caption{(a) Illustration of transformations; (b) Virtual rendering example of the large needle driver and its keypoint locations; (c) Extracted gradient orientations from virtual rendering. The orientations are quantised and colour-coded as shown in the pie chart.}
\label{fig:trans}
\end{figure}
\section{Methods}
Our proposed framework includes three main components. The first component is a virtual tool renderer that generates part-based templates online. After template matching, the second component performs verification to extract the inlier 2D detections. These 2D detections are finally fused with kinematic data for 3D tool pose estimation. Our framework is implemented on the da Vinci\textsuperscript{\textregistered} robot. The robot kinematics are retrieved using the da Vinci\textsuperscript{\textregistered} Research Kit (dVRK) \cite{Kazanzides2014}.

\subsection{Part-based Online Templates for Tool Detection} \label{sec:detect}
In this work, to deal with the changing pose of articulated surgical tools, the tool detection has been performed by matching individual parts of the tools, rather than the entire instrument, similar to \cite{Sznitman2014}. To avoid the limitations of offline training, we propose to generate the part models on-the-fly such that the changing appearance of tool parts can be dynamically adapted. 

To generate the part-based models online, the CAD model of the tool and the robot kinematics have been used to render the tool in a virtual environment. The pose of a tool in the robot base frame $B$ can be denoted as the transformation $\mathbf{T}_{E}^{B}$, where $E$ is the end-effector coordinate frame shown in Fig.\ref{fig:trans}(a). $\mathbf{T}_{E}^{B}$ can be retrieved from dVRK (kinematics) to provide the 3D coordinates of the tool in $B$. Thus, to set the virtual view to be the same as the laparoscopic view, a standard hand-eye calibration \cite{Tsai1989} is used to estimate the transformation $\mathbf{T}_{B}^{C}$ from $B$ to the camera coordinate frame $C$. However, errors in the calibration can affect the accuracy of $\mathbf{T}_{B}^{C}$, resulting in a 3D pose offset between the virtual tool and the real tool in $C$. In this regard, we represent the transformation found from the calibration as $\mathbf{T}_{B}^{C^{-}}$, where $C^{-}$ is the camera coordinate frame that includes the accumulated calibration errors. Therefore, a correction transformation denoted as $\mathbf{T}_{C^{-}}^{C}$ can be introduced to compensate for the calibration errors.
\begin{figure*}[t]
\centering
\includegraphics[width=\linewidth]{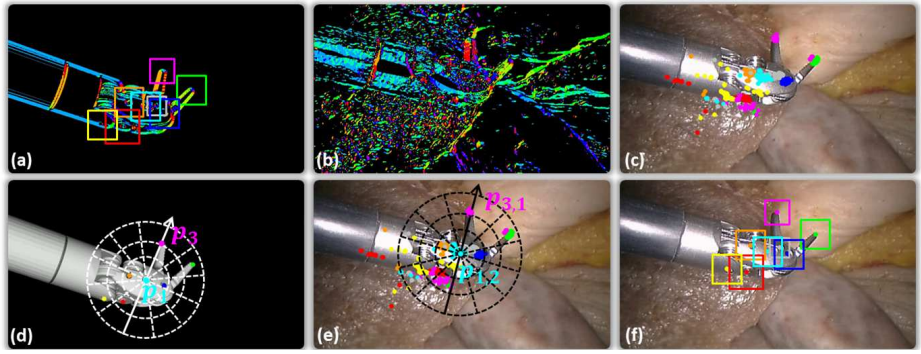}
\caption{(a) An example of part-based templates; (b) Quantised gradient orientations from a camera image; (c) Part-based template matching results of tool parts; (d) and (e) Geometrical context verification; (f) Inlier detections obtained after verification.}
\label{fig:verify}
\end{figure*}

In this work, we have defined $n$=$14$ keypoints $\mathbf{P}^{B}=\left\lbrace p_{i}^{B} \right\rbrace_{i=1}^{n}$ on the tool, and the large needle driver is taken as an example. The keypoints include the points shown in Fig.\ref{fig:trans}(b) and those on the symmetric side of the tool. These keypoints represent the skeleton of the tool, which also apply to other da Vinci\textsuperscript{\textregistered} tools. At time $t$, an image $I_{t}$ can be obtained from the laparoscopic camera. The keypoints can be projected in $I_{t}$ with the camera intrinsic matrix $\mathbf{K}$ via
\begin{equation} \label{eq:func_P}
\mathbf{P}_{t}^{I} = \frac{1}{s} \mathbf{K} \mathbf{T}_{C^{-}}^{C} \mathbf{T}_{B}^{C^{-}} \mathbf{P}_{t}^{B}.
\end{equation}
Here, $s$ is the scaling factor that normalises the depth to the image plane. 

To represent the appearance of the tool parts, the Quantised Gradient Orientations (QGO) approach \cite{Hinterstoisser2012} has been used (see Fig.\ref{fig:trans}(c)). Bounding boxes are created to represent part-based models and centred at the keypoints in the virtual view (see Fig.\ref{fig:verify}(a)). The box size for each part is adjusted according to the $z$ coordinate of the keypoint with respect to the virtual camera centre. QGO templates are then extracted inside these bounding boxes. As QGO represents the contour information of the tool, it is robust to cluttered scenes and illumination changes. In addition, a QGO template is represented as a binary code by quantisation, thus template matching can be performed efficiently.

Note that not all of the defined parts are visible in the virtual view, as some of them may be occluded. Therefore, the templates are only extracted for those $m$ parts that facing the camera. To find the correspondences of the tool parts between the virtual and real images, QGO is also computed on the real image (see Fig.\ref{fig:verify}(b)) and template matching is then performed for each part via sliding windows. Exemplar template matching results are shown in Fig.\ref{fig:verify}(c).
\subsection{Tool Part Verification via 2D Geometrical Context} \label{sec:verify}
To further extract the best location estimates of the tool parts, a consensus-based verification approach \cite{Ye2016} is included. This approach analyses the geometrical context of the correspondences in a PROgressive SAmple Consensus (PROSAC) scheme \cite{Chum2005}. For the visible keypoints $\left\lbrace p_{i}\right\rbrace_{i=1}^{m}$ in the virtual view, we denote their 2D correspondences in the real camera image as $\left\lbrace  p_{i,j} \right\rbrace_{i=1,j=1}^{m,k}$, where $\left\lbrace p_{i,j}\right\rbrace_{j=1}^{k}$ represent the top $k$ correspondences of $p_{i}$ sorted by QGO similarities.

For each iteration in PROSAC, we select two point pairs from $\left\lbrace  p_{i,j} \right\rbrace_{i=1,j=1}^{m,k}$ in a sorted descending order. These two pairs represent the correspondences for two different parts, e.g., pair of $p_{1}$ and $p_{1,2}$, and pair of $p_{3}$ and $p_{3,1}$. The two pairs are then used to verify the geometrical context of the tool parts. As shown in Fig.\ref{fig:verify}(d) and (e), we use two polar grids to indicate the geometrical context of the virtual view and the camera image. The origins of the grids are defined as $p_{1}$ and $p_{1,2}$, respectively. The major axis of the grids can be defined as the vectors from $p_{1}$ to $p_{3}$ and $p_{1,2}$ to $p_{3,1}$, respectively. The scale difference between the two grids is found by comparing $d\left(p_{1}, p_{3}\right)$ and $d\left(p_{1,2}, p_{3,1}\right)$, where $d\left(\cdot, \cdot \right)$ is the euclidean distance. We can then define the angular and radial bin sizes as $30$ degrees and $10$ pixels (allowing moderate out-of-plane rotation), respectively. With these, two polar grids can be created and placed on the virtual and camera images. A point pair is determined as an inlier if the two points are located in the same zone in the polar grids. Therefore, if the number of inliers is larger than a predefined value, the geometrical context of the tools in the virtual and the real camera images are considered as matched. Otherwise, the above verification is repeated until it reaches the maximum number (100) of iterations. After verification, the inlier point matches are used to estimate the correction transformation $\mathbf{T}_{C^-}^{C}$.    
\subsection{From 2D to 3D Tool Pose Estimation} \label{sec:posest}
We now describe how to combine the 2D detections with 3D kinematic data to estimate $\mathbf{T}_{C^{-}}^{C}$. Here the transformation matrix is represented as a vector of rotation angles and translations along each axis: $\mathbf{x} = \left[ \theta_x, \theta_y, \theta_z, r_x, r_y, r_z \right]^T$. We denote the $n$ observations (corresponding to the tool parts defined in Section \ref{sec:detect}) as $\mathbf{z} = \left[ u_1, v_1, \ldots, u_n, v_n \right]^T$, where $u$ and $v$ are their locations in the camera image. To estimate $\mathbf{x}$ on-the-fly, the EKF has been adopted to find $\mathbf{x}_t$ given the observations $\mathbf{z}_t$ at time $t$. The process model is defined as $\mathbf{x}_{t} = \mathbf{I} \mathbf{x}_{t-1} + \mathbf{w}_{t},$ where $\mathbf{w}_t$ is the process noise at time $t$, and $\mathbf{I}$ is the transition function defined as the identity matrix. The measurement model is defined as $\mathbf{z}_t = \mathbf{h}(\mathbf{x}_{t}) + \mathbf{v}_t$, with $\mathbf{v}_t$ being the noise. $\mathbf{h}(\cdot)$ is the nonlinear function with respect to $\left[ \theta_x, \theta_y, \theta_z, r_x, r_y, r_z \right]^T$: 
\begin{equation} \label{eq:func_h}
\mathbf{h}(\mathbf{x}_{t}) = \frac{1}{s} \mathbf{K} \mathbf{f}(\mathbf{x}_t) \mathbf{T}_{B}^{C^{-}} \mathbf{P}_{t}^{B},
\end{equation}
which is derived according to Eq.\ref{eq:func_P}. Note here, $\mathbf{f}(\cdot)$ is the function that composes the euler angles and translation (in $\mathbf{x}_t$) into the $4\times4$ homogenous transformation matrix $\mathbf{T}_{C^{-}}^{C}$. As Eq.\ref{eq:func_h} is a nonlinear function, we derive the Jacobian matrix $\mathbf{J}$ of $\mathbf{h}(\cdot)$ regarding each element in $\mathbf{x}_t$.

For iteration $t$, the predicted state $\mathbf{\mathbf{x}^{-}_{t}}$ is calculated and used to predict the measurement $\mathbf{\mathbf{z}^{-}_{t}}$, and also to calculate $\mathbf{J}_t$. In addition, $\mathbf{z}_t$ is obtained from the inlier detections (Section \ref{sec:verify}), which is used, along with $\mathbf{J}_t$ and $\mathbf{\mathbf{x}^{-}_{t}}$, to derive the corrected state $\mathbf{\mathbf{x}^{+}_{t}}$ which contains the corrected angles and translations. These are finally used to compose the transformation $\mathbf{T}_{C^{-}}^{C}$ at time $t$, and thus the 3D pose of the tool in $C$ is obtained as $\mathbf{T}_{E}^{C} = \mathbf{T}_{C^{-}}^{C}\mathbf{T}_{B}^{C^-}\mathbf{T}_{E}^{B}$. Note that if no 2D detections are available at time $t$, the previous $\mathbf{T}_{C^{-}}^{C}$ is then used.

At the beginning of the tracking process, an estimate $^{0}\mathbf{T}_{C^{-}}^{C}$ is required to initialise EKF, and correct the virtual view to be as close as possible to the real view. Therefore, template matching is performed in multiple scales and rotations for initialisation, however, only one template is needed for matching of each tool part after initialisation. The Efficient Perspective-n-Points (EPnP) algorithm \cite{Lepetit2008} is applied to estimate $^{0}\mathbf{T}_{C^{-}}^{C}$ based on the 2D-3D correspondences of the tool parts matched between the virtual and real views and their 3D positions from kinematic data.

The proposed framework can be easily extended to track multiple tools. This only requires to generate part-based templates for all the tools in the same graphic rendering and follow the proposed framework. As template matching is performed in binarised templates, the computational speed is not deteriorated. 
\section{Results}
The proposed framework has been implemented on an HP workstation with an Intel Xeon E5-2643v3 CPU. Stereo videos are captured at $25$Hz. In our \texttt{C++} implementation, we have separated the part-based rendering and image processing into two CPU running threads, enabling our framework to be real-time. The rendering part is implemented based on VTK and OpenGL, of which the speed is fixed as $25$Hz. As our framework only requires monocular images for 3D pose estimation, only the images from the left camera were processed. For image size 720x576, the processing speed is $\approx29$Hz (without any GPU programming). The threshold of the inlier number in the geometrical context verification is empirically defined as $4$. For initialisation, template matching is performed with additional scale ratios of $0.8$ and $1.2$, and rotations of $\pm15$ degrees, which does not deteriorate the run-time speed due to template binarisation. Our method was compared to the tracking approaches for articulated tools including \cite{Sznitman2014} and \cite{Reiter2014}.
\begin{figure*}[t]
\centering
\includegraphics[width=\linewidth]{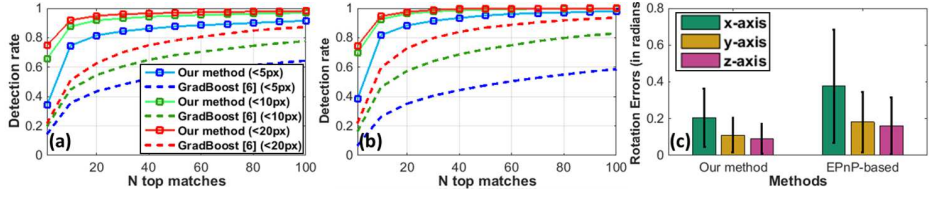}
\caption{(a) and (b) Detection rate results of our online template matching and GradBoost \cite{Sznitman2014} on two single-tool tracking sequences (see supplementary videos); (c) Overall rotation angle errors (mean$\pm$std) along each axis on Seqs.1-6.}
\label{fig:detectrate}
\end{figure*}
\begin{table}[t]
\footnotesize 
\centering
\renewcommand{\arraystretch}{1.0}
\caption{ Translation and rotation errors (mean$\pm$std) on Seqs.1-6. Tracking accuracies with run-time speed in Hz (in brackets) compared to \cite{Reiter2014} on their dataset (Seqs.7-12).}
\scalebox{0.9}{
\begin{tabular}{ccccc|ccc}
\hline
 & \multicolumn{4}{c|}{\textbf{3D Pose Error}} &  &\multicolumn{2}{c}{\textbf{Tracking Accuracy}} \\ \hline
\multicolumn{1}{c|}{\multirow{2}{*}{Seq.}} & \multicolumn{2}{c}{\textbf{Our method}} & \multicolumn{2}{c|}{EPnP-based}  & \multicolumn{1}{c|}{\multirow{2}{*}{Seq.}} &  \multirow{2}{*}{\textbf{Our method}} & \multirow{2}{*}{\cite{Reiter2014}} \\ 
\multicolumn{1}{c|}{} & \multicolumn{1}{c}{Trans.(mm)} & \multicolumn{1}{c}{Rot.(rads.)} & \multicolumn{1}{c}{Trans.(mm)} & Rot.(rads.) & \multicolumn{1}{c|}{} &  \multicolumn{1}{c}{} &   \\ \hline
\multicolumn{1}{c|}{1} & $\mathbf{1.31\pm1.15}$ & $\mathbf{0.11\pm0.08}$ & $3.10\pm3.89$ & $0.12\pm0.09$ & \multicolumn{1}{c|}{7} &  \textbf{97.79\%(27)} & 97.12\%(1) \\ \hline
\multicolumn{1}{c|}{2} & $\mathbf{1.50\pm1.12}$ & $\mathbf{0.12\pm0.07}$ & $6.69\pm8.33$ & $0.24\pm0.19$ & \multicolumn{1}{c|}{8} &\textbf{99.45\%(27)} & 96.88\%(1) \\ \hline
\multicolumn{1}{c|}{3} & $\mathbf{3.14\pm1.96}$ & $\mathbf{0.12\pm0.08}$ & $8.03\pm8.46$ & $0.23\pm0.20$ & \multicolumn{1}{c|}{9} & \textbf{99.25\%(28)} & 98.04\%(1) \\ \hline
\multicolumn{1}{c|}{4} & $\mathbf{4.04\pm2.21}$ & $\mathbf{0.19\pm0.15}$ & $5.02\pm5.41$ & $0.29\pm0.18$  & \multicolumn{1}{c|}{10} & 96.84\%\textbf{(28)} & \textbf{97.75\%}(1) \\ \hline
\multicolumn{1}{c|}{5} & $\mathbf{3.07\pm2.02}$ & $\mathbf{0.14\pm0.11}$ & $5.47\pm5.63$ & $0.26\pm0.20$  & \multicolumn{1}{c|}{11} & 96.57\%\textbf{(36)} & \textbf{98.76\%}(2) \\ \hline
\multicolumn{1}{c|}{6} & $\mathbf{3.24\pm2.70}$ & $\mathbf{0.12\pm0.05}$ & $4.03\pm3.87$ & $0.23\pm0.13$  & \multicolumn{1}{c|}{12} & \textbf{98.70\%(25)} & 97.25\%(1) \\ \hline
\multicolumn{1}{c|}{Overall} & $\mathbf{2.83\pm2.19}$ & $\mathbf{0.13\pm0.10}$ & $5.51\pm6.45$ & $0.24\pm0.18$ & \multicolumn{1}{c|}{Overall} &  \textbf{97.83\%} & 97.81\%  \\ \hline
\end{tabular}
}
\label{tab:errors}
\end{table}

For demonstrating the effectiveness of the online part-based templates for tool detection, we have compared our approach to the method proposed in \cite{Sznitman2014}, which is based on boosted trees for 2D tool part detection. For ease of training data generation, a subset of the tool parts was evaluated in this comparison, namely the front pin, logo, and rear pin. The classifier was trained with $6000$ samples for each part. Since \cite{Sznitman2014} applies to single tool tracking only, the trained classifier along with our approach were tested on two single-tool sequences (1677 and 1732 images), where ground truth data was manually labelled. A part detection is determined to be correct if the distance of its centre and ground truth is smaller than a threshold. To evaluate the results with different accuracy requirements, the threshold was therefore sequentially set to 5, 10, and 20 pixels. The detection rates of the methods were calculated among the top N detections. As shown in Fig.\ref{fig:detectrate}(a-b) our method significantly outperforms \cite{Sznitman2014} in all accuracy requirements. This is because our templates are generated adaptively online.
\begin{figure*}[t]
\centering
\includegraphics[width=\linewidth]{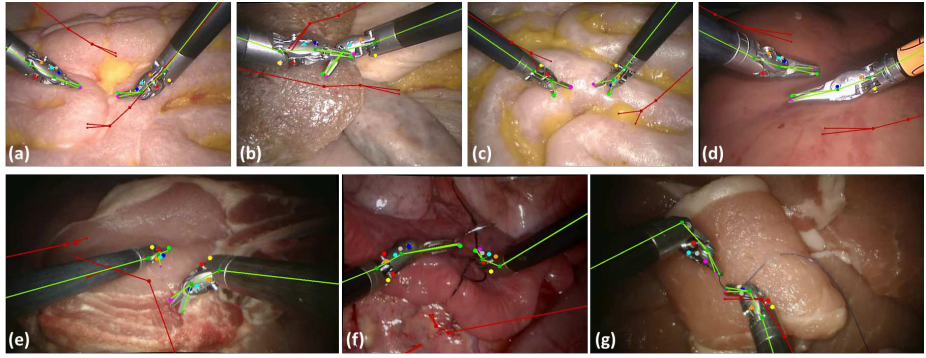}
\caption{Qualitative results. (a-c) phantom data (Seqs.1-3); (d) \emph{ex vivo} ovine data (Seq.4); (e) and (g) \emph{ex vivo} porcine data (Seqs.9 and 12); (f) \emph{in vivo} porcine data (Seq.11). Red lines indicate the tool kinematics, and green lines indicate the tracking results of our framework with 2D detections in coloured dots.}
\label{fig:results}
\end{figure*}

To validate the accuracy of the 3D pose estimation, we manually labelled the centre locations of the tool parts on both left and right camera images on phantom (Seqs.1-3) and \emph{ex vivo} (Seqs.4-6) video data to generate the 3D ground truth. The tool pose errors are then obtained as the relative pose between the estimated pose and the ground truth. Our approach was also compared to the 3D poses estimated performing EPnP for every image where the tool parts are detected. However, EPnP generated unstable results and had inferior performance to our approach as shown in Table \ref{tab:errors} and Fig.\ref{fig:detectrate}(c).

We have also compared our framework to the method proposed in \cite{Reiter2014}. As their code is not publically available, we ran our framework on the same \emph{ex vivo} (Seqs.7-10,12) and \emph{in vivo} data (Seq.11) used in \cite{Reiter2014}. Example results are shown in Fig.\ref{fig:results}(e-g). For achieving a fair comparison, we have evaluated the tracking accuracy as explained in their work, and presented both our results and theirs reported in the paper in Table \ref{tab:errors}. Although our framework achieved slightly better accuracies than their approach, our processing speed is significantly faster, ranging from $25\textup{--}36$Hz, while theirs is approximately $1\textup{--}2$Hz as reported in \cite{Reiter2014}. As shown in Figs.\ref{fig:results}(b) and (d), our proposed method is robust to occlusion due to tool intersections and specularities, thanks to the fusion of 2D part detections and kinematics. In addition, our framework is able to provide accurate tracking even when $\mathbf{T}_{B}^{C^{-}}$ becomes invalid after the laparoscope has moved (Fig.\ref{fig:results}(c), Seq.3). This is because $\mathbf{T}_{C^{-}}^{C}$ is estimated online using the 2D part detections. All the processed videos are available via \url{https://youtu.be/oqw_9Xp_qsw}.

\section{Conclusions}
In this paper, we have proposed a real-time framework for 3D tracking of articulated tools in robotic surgery. Online part-based templates are generated using the tool CAD models and robot kinematics, such that efficient 2D detection can then be performed in the camera image. For rejecting outliers, a robust verification method based on 2D geometrical context is included. The inlier 2D detections are finally fused with robot kinematics for 3D pose estimation. Our framework can run in real-time for multi-tool tracking, thus can be used for imposing dynamic active constraints and motion analysis. The results on phantom, \emph{ex vivo} and \emph{in vivo} experiments demonstrate that our approach can achieve accurate 3D tracking, and outperform the current state-of-the-art.
\subsubsection*{Acknowledgements.} We would like to thank Simon DiMaio from Intuitive Surgical for providing the tool CAD models, and Austin Reiter from Johns Hopkins University for assisting method comparisons. 

\bibliographystyle{splncs03}
\bibliography{miccai}

\begin{thebibliography}{10}
\providecommand{\url}[1]{\texttt{#1}}
\providecommand{\urlprefix}{URL }

\bibitem{Sznitman2012}
Sznitman, R., Ali, K., Richa, R., Taylor, R.H., Hager, G.D., Fua, P.:
  Data-driven visual tracking in retinal microsurgery. In: Ayache, N.,
  Delingette, H., Golland, P., Mori, K. (eds.) MICCAI 2012, vol. 7511, pp.
  568--575. Springer, Heidelberg (2012)

\bibitem{Wolf2011}
Wolf, R., Duchateau, J., Cinquin, P., Voros, S.: 3d tracking of laparoscopic
  instruments using statistical and geometric modeling. In: Fichtinger, G.,
  Martel, A., Peters, T. (eds.) MICCAI 2011, vol. 6891, pp. 203--210. Springer,
  Heidelberg (2011)

\bibitem{Allan2015}
Allan, M., Chang, P.L., Ourselin, S., Hawkes, D.J., Sridhar, A., Kelly, J.,
  Stoyanov, D.: Image based surgical instrument pose estimation with
  multi-class labelling and optical flow. In: Navab, N., Hornegger, J., Wells,
  M.W., Frangi, F.A. (eds.) MICCAI 2015, vol. 9349, pp. 331--338. Springer,
  Heidelberg (2015)

\bibitem{Pezzementi2009}
Pezzementi, Z., Voros, S., Hager, G.: Articulated object tracking by rendering
  consistent appearance parts. In: ICRA. pp. 3940--3947 (2009)

\bibitem{Reiter2012}
Reiter, A., Allen, P.K., Zhao, T.: Articulated surgical tool detection using
  virtually-rendered templates. In: CARS (2012)

\bibitem{Sznitman2014}
Sznitman, R., Becker, C., Fua, P.: Fast part-based classification for
  instrument detection in minimally invasive surgery. In: Golland, P., Hata,
  N., Barillot, C., Hornegger, J., Howe, R. (eds.) MICCAI 2014, vol. 8674, pp.
  692--699. Springer, Heidelberg (2014)

\bibitem{Rieke2015}
Rieke, N., Tan, D.J., Alsheakhali, M., Tombari, F., di~San~Filippo, C.A.,
  Belagiannis, V., Eslami, A., Navab, N.: Surgical tool tracking and pose
  estimation in retinal microsurgery. In: Navab, N., Hornegger, J., Wells,
  M.W., Frangi, F.A. (eds.) MICCAI 2015, vol. 9349, pp. 266--273. Springer,
  Heidelberg (2015)

\bibitem{Reiter2014}
Reiter, A., Allen, P.K., Zhao, T.: Appearance learning for 3d tracking of
  robotic surgical tools. Int. J. Rob. Res.  33(2),  342--356 (2014)

\bibitem{Kazanzides2014}
Kazanzides, P., Chen, Z., Deguet, A., Fischer, G., Taylor, R., DiMaio, S.: An
  open-source research kit for the da vinci$^{\circledR}$ surgical system. In:
  ICRA. pp. 6434--6439 (2014)

\bibitem{Tsai1989}
Tsai, R., Lenz, R.: A new technique for fully autonomous and efficient 3d
  robotics hand/eye calibration. IEEE Trans. Rob. Autom.  5(3),  345--358
  (1989)

\bibitem{Hinterstoisser2012}
Hinterstoisser, S., Cagniart, C., Ilic, S., Sturm, P., Navab, N., Fua, P.,
  Lepetit, V.: Gradient response maps for real-time detection of textureless
  objects. IEEE Trans. Pattern Anal. Mach. Intell.  34(5),  876--888 (2012)

\bibitem{Ye2016}
Ye, M., Giannarou, S., Meining, A., Yang, G.Z.: Online tracking and retargeting
  with applications to optical biopsy in gastrointestinal endoscopic
  examinations. Med. Image Anal.  30,  144--157 (2016)

\bibitem{Chum2005}
Chum, O., Matas, J.: Matching with prosac - progressive sample consensus. In:
  CVPR. vol.~1, pp. 220--226 (2005)

\bibitem{Lepetit2008}
Lepetit, V., Moreno-Noguer, F., Fua, P.: Epnp: An accurate o(n) solution to the
  pnp problem. Int. J. Comput. Vision  81(2),  155--166 (2008)

\end{thebibliography}
\end{document}